\documentclass{article}
\usepackage{microtype}
\usepackage{graphicx}
\usepackage{subfigure}
\usepackage{booktabs} % for professional tables

\usepackage{hyperref}

\usepackage[accepted]{icml2023}

\usepackage{amsmath}
\usepackage{amssymb}
\usepackage{mathtools}
\usepackage{amsthm}

\usepackage[capitalize,noabbrev]{cleveref}

\theoremstyle{plain}

\theoremstyle{definition}

\theoremstyle{remark}

\usepackage[textsize=tiny]{todonotes}

\icmltitlerunning{Sampling-based Path Planning Algorithms: A Survey}

\begin{document}

\twocolumn[
\icmltitle{Sampling-based Path Planning Algorithms: A Survey}



\icmlsetsymbol{equal}{*}

\begin{icmlauthorlist}
\icmlauthor{Alka Choudhary}{WPI}
\end{icmlauthorlist}

\icmlaffiliation{WPI}{WPI, Worcester, USA}

\icmlcorrespondingauthor{Alka Choudhary}{mnnitalka@gmail.com}

\icmlkeywords{Motion Planning, Sampling-based Path-planning Algorithms, Autonomous Driving}

\vskip 0.3in
]



\printAffiliationsAndNotice{} % otherwise use the standard text.

\begin{abstract}
Path planning is a classic problem for autonomous robots. To ensure safe and efficient point-to-point navigation an appropriate algorithm should be chosen keeping the robot's dimensions and its classification in mind. Autonomous robots use path-planning algorithms to safely navigate a dynamic, dense, and unknown environment. A few metrics for path planning algorithms to be taken into account are safety, efficiency, lowest-cost path generation, and obstacle avoidance. Before path planning can take place we need map representation which can be discretized or open configuration space. Discretized configuration space provides node/connectivity information from one point to another. While in open/free configuration space it is up to the algorithm to create a list of nodes and then find a feasible path. Both types of maps are populated by obstacle positions using perception obstacle detection techniques to represent current obstacles from the perspective of the robot. For open configuration spaces, sampling-based planning algorithms are used. This paper aims to explore various types of Sampling-based path-planning algorithms such as Probabilistic RoadMap (PRM), and Rapidly-exploring Random Trees (RRT). These two algorithms also have optimized versions - PRM* and RRT* and this paper discusses how that optimization is achieved and is beneficial.
\end{abstract}

\section{Introduction}
The application areas of autonomous robots have seen exponential growth in the last decade. Today they are used in distribution centers, security,  healthcare, hospitality, grocery stores, last-mile delivery, self-driving cars, etc. Each market and industry is heading toward automation and eventually will start using autonomous robots. Robot motion is a complex problem to solve given the robot's dynamics and constantly changing environment. Navigation requires modeling of the environment \cite{widjaja2022machine} and localization to understand the robot's current position within the environment. First, it detects obstacles \cite{park20233m3d} and then creates an obstacle-free path to the goal providing control inputs on how to reach there. Path planning plays a key role to find feasible trajectories to reach the goal. 

\begin{figure}[ht]
\vskip 0.2in
\begin{center}
\centerline{\includegraphics[width=\columnwidth]{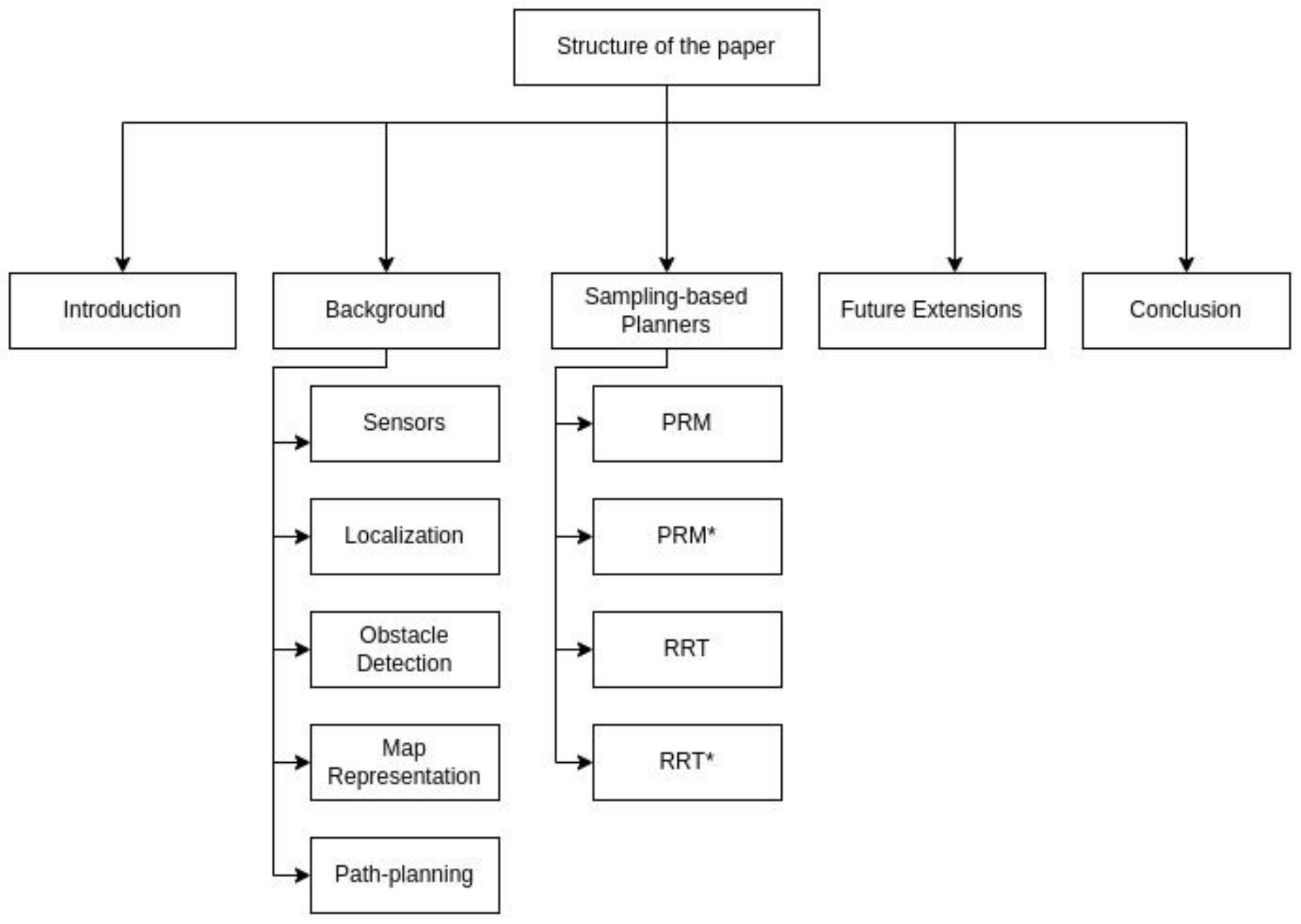}}
\caption{Structure of the paper}
\label{paper-structure}
\end{center}
\vskip -0.2in
\end{figure}

Sampling-based motion planning is suitable for open configuration spaces which are large in their size. The aim of these algorithms is to find an obstacle-free path in minimum time which come with the caveat of not guaranteeing the optimality of the solution. This paper explores various Sampling-based path-planning algorithms that evolved in the last few decades. These algorithms can be categorized into two major categories - Probabilistic RoadMaps (PRMs) and Rapidly-exploring Random Trees(RRT). This survey paper provides an overall deep understanding of these two algorithms and optimization techniques that were used to improve performance in algorithms like - PRM*, and RRT*. The structure of this paper is as shown in \cref{paper-structure}.

\section{Background}
\label{background_section}
\begin{figure}[ht]
\vskip 0.2in
\begin{center}
\centerline{\includegraphics[width=\columnwidth]{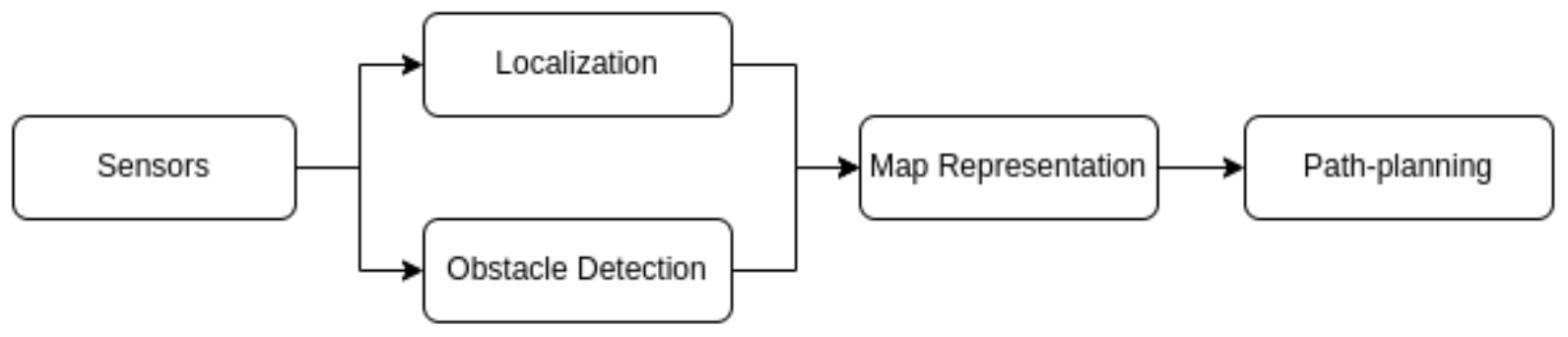}}
\caption{Flow chart of Robot's operation and data flow into Path-planners }
\label{robot-operation}
\end{center}
\vskip -0.2in
\end{figure}

Before we dive into sampling-based path-planning algorithms it is important to understand how path planning fits into the context of end-to-end robot operations. As shown in the \cref{robot-operation} there are a few key components Sensors, Localization, Obstacle detection, Map representation, and Path-planning. Path-planner gets information about the surroundings and current location before it can start planning for the goal. There are various other components for overall robot operation but for setting the context of this paper we will be discussing on which Path-planner depends.

\subsection{Sensors}
Humans navigate in the world by sensing the environment which makes them aware of their own location and also moving parts of the external environment. For a robot to perform any task it also needs sensory data to perceive the environment it is operating in. The next question comes to what type of sensors are available and what are their use cases. These sensors can further be sub-categorized into the 2D Camera, 3D Camera, Topographic Lidar, and Bathymetric Lidar. The most common sensors used are - Camera, Lidar, Radar, IMU, etc. In autonomous driving multi-camera setups combined with Lidars are mostly used \cite{singh2023surround}. Overall sensor setup gathers raw information about all the objects around it and then it passes it to the localization and obstacle detection (perception) unit.

\subsection{Localization}
The localization sub-system tells the robot its current location with respect to the global frame. For e.g. how Google Maps can give information about your current location on the map. Why is it important? Before planning the goal robot needs to get the right start location which comes from localization. The reliability of the localization unit is the most important factor - how confident is it this system that the robot is at X position right now? Localization is done by using sensor input, previously logged locations, and fixed maps. These fixed maps generally are generally generated by mapping static components of the environment for e.g. in a city mapping building, traffic signals, etc.

\subsection{Obstacle Detection}
All the sensors provide raw data which to the human eye doesn't make any sense but it contains precise locations of an object with respect to the robot. This sensor data is processed by the Obstacle Detection / Perception unit to map these objects on the map. Some of these sensors when used standalone are not sufficient to make a confident decision where an object is seen. A multi-sensor setup provides various types of raw data to come to a more confident decision. Recently in the autonomous vehicle industry, the focus has been on detecting 3D obstacles to provide a better map representation of objects around a card. These are the various state-of-art techniques to detect 3D obstacles like vision-radar-based fusion \cite{singh2023vision}, surround-view vision-based detection \cite{singh2023surround}, and transformer-based sensor fusion \cite{singh2023transformer} etc.

\subsection{Map Representation}
Both localization and obstacle detection is required before finalizing the map representation to be sent to the path planners. Map representation sub-system consists of these two components:
\begin{itemize}
\item \textbf{Static Map} is created by mapping static components of the environment. These static components do not move with time for e.g. a building on the side of a road.
\item \textbf{Dynamic obstacle detection updates} are done when the perception unit provides information about an obstacle with respect to the current robot's position provided by the localization unit.
\end{itemize}

Map/configuration space consists of all the positions/configurations that the robot can reach. These maps can be represented in the form of costmaps or configuration space. Every point in the map have a cost for e.g. area within an obstacle is assigned a lethal cost so that planner do not plan through it.

\subsection{Path-planning}
Path planning comes into the category of non-deterministic polynomial-time (NP) hard problem \cite{4666104} to find the path from the start to the goal location. As we put robots in a dynamic environment complexities increase, and the complexity of the algorithm also increases with an increase in the degrees of freedom of the robot \cite{vehicles3030027}. Navigation through a complex and dynamic environment poses challenges to generating a path that is safe and efficient.

Path-planner takes costmap, start, and goal location as input to produce a path. Apart from cost areas, these maps can be discretized with added node connectivity. Grid-based planning like A*, Dijkstra, etc use discretized maps and generate an optimal low-cost path to the goal. Limitations of these types of algorithms arise in bigger maps where discretization is not possible and then sampling-based algorithms are the better choice. There are two criteria on which planning algorithms are rated:
\begin{itemize}
    \item \textbf{Feasible:} A plan is guaranteed generated to the goal if it is possible to reach to that point, efficiency is not a concern here.
    \item \textbf{Optimal:} Optimizing the performance of the planner in addition to finding the path to the goal.
\end{itemize}

In simpler terms, a path planner generates a plan which is a sequence of actions taken to reach the goal state. Robots today operate in dynamically changing environments so these planners also need to accommodate those functions of the state. Another way to categorize the planner is - Global vs Local planner. Global planners are used to generating a path from start to goal working with global costmaps while local planner work with smaller local costmaps and are used in trajectory generation and following.

\section{Sampling-based Path Planning}
Sampling-based path planner randomly connects points in the state space and constructs a graph to create obstacle-free paths \cite{doi:10.1177/0278364911406761}. These algorithm doesn't require exploring the full configuration space so they are faster and more efficient. The number of iterations to generate the graph connectivity can be set by the user which will dictate the optimality of the path that it finds\cite{motion_planning_survey_2}. These type of algorithms presents a significant issue while traversing tight spaces as it is difficult to find the connectivity through narrow spaces via random sampling. Following are the different types of Sampling-based Path Planners.

Probabilistic Roadmap (PRM), and Radpid-exploring Random Trees (RRT) are the two most discussed algorithms in Sampling-based path planners. Differentiation in these comes from the way they connect points to create the graph. PRM* and RRT* are the optimized versions of these algorithms which we will discuss further in this paper.

\subsection{PRM}
PRM (Probabilistic Roadmap) is one of the initial sampling-based path planners. PRM is a graph containing nodes and edges in a map consisting of obstacles and obstacle-free areas. First, it generates randomly sampled nodes in the configuration space and then connects the current node to its neighboring nodes if the edge is in an obstacle-free area. PRM also takes the radius as an input to determine which random neighbors are calculated and going to be connected. To generate random sampling nodes this paper \cite{doi:10.1177/0278364911406761} describes a variety of methods. Biases from the random point generator also affect the results of PRM graphs.

After generating random neighboring nodes within a fixed radius, PRM divides them into smaller connected graphs/clusters. These clusters are circular with a radius as described by the user. Every node can belong to multiple connected clusters. Neighboring nodes are sorted by a metric e.g. increasing distance from the current node. Each neighbor node is checked if it belongs to the same connected cluster as the current node is in, if not then it is added to the same cluster.

Parsed radius will determine the performance of PRM generated graph, the bigger the radius more neighbors will be generated and determined for their cluster association. We can also parse a parameter total number of nodes to be generated by the PRM algorithm the more this number is it will increase time PRM \cref{PRM-Algorithm} takes to create the graph. But if this number is too low, it can generate a fragmented graph. The limitations of PRM come in place in the obstacle-dense regions and present the issue of fractured graphs. Looking for the shortest path is also challenging for the resulting sparse graph \cite{elbanhawi2014sampling}.

\begin{algorithm}[tb]
\caption{PRM Algorithm}
\label{PRM-Algorithm}
\begin{algorithmic}
    \STATE {\bfseries Input:} $graph, iteration\_limit, radius$
   \REPEAT
   \STATE Initialize $i = 0$
   \STATE $new\_node = random\_sampler()$
   \STATE $nearest\_n = nearest(graph, new\_node, radius)$
   \STATE $nearest\_n = sort(nearest\_n)$
   \FOR{$i=1$ {\bfseries to} $number\_of\_nearest\_nodes$}
        \STATE $edge\_link = createEdge(new\_node,nearest[i])$
        \STATE $j = edgeNotInObstacleReigon(edge\_link)$
        \STATE $k = in\_same\_cluster(new\_node,nearest[i])$
        \IF{$j\ and\ k$}
            \STATE $addToGraph(edge\_link)$
            \STATE $nearest\_n[i].cluster = new\_node.cluster$
        \ENDIF
   \ENDFOR
   \STATE $i = i + 1$
   \UNTIL{$i < iteration\_limit$}
\end{algorithmic}
\end{algorithm}

\subsection{PRM*}
PRM* is an optimized version of the PRM (Probabilistic Roadmap) algorithm. It provides two modifications to the default PRM algorithm. In PRM* the concept of a fixed radius is removed instead it is a dynamic variable that keeps changing based on the number of nodes that have already been generated. Here is the function:
\cite{doi:10.1177/0278364911406761} to calculate this radius
\begin{equation}
R_{PRM*} = \gamma PRM* \left(\frac{\log(n)}{n}\right) ^ {\!1/d}
\end{equation}
\begin{equation}
\gamma PRM* > {2}\left( 1 + \frac{1}{d}\right) ^ {\!1/d} (\mu (X(free))/ \zeta_d) ^ {\!1/d}
\end{equation}

Radius $R_{PRM^*}$ depends upon the number of nodes ${n}$, the constant characteristic of the configuration space ${\gamma PRM*}$, and the number of dimensions ${d}$. Calculation of ${\gamma}$ constant is a factor of ${\mu (Xfree)}$, free space available for path planning, and ${\zeta_d}$ is the unit sphere volume. ${\log(n)/n}$ term keep decreasing as we add more nodes which in effect causes a decrease in radius to calculate neighboring nodes.
\\ This provides a significant improvement in cluster creation. We can see more straight roads after a path is generated. Think about it as driving in and outside the city, early generated nodes can be considered as far away cities, and edges to connect them are straight highways. As we start to grow in the number of nodes, the radius decrease and the nodes created are closer and can be considered like a neighborhood of a city and edges becomes smaller roads within the city.
\\ Next optimization made to ${PRM*}$ is to get rid of the clustering system. Instead, any node within a radius become connected. It helps to reduce the complexity of \cref{PRM*-Algorithm}. PRM* generates a graph that is very dense. A dense graph provides a smoother path while ${PRM}$ presented a zigzagged path.

\begin{algorithm}[tb]
\caption{PRM* Algorithm}
\label{PRM*-Algorithm}
\begin{algorithmic}
    \STATE {\bfseries Input:} $graph, iteration\_limit$
   \REPEAT
   \STATE Initialize $i = 0$
   \STATE $new\_node = random\_sampler()$
   \STATE $R_{PRM*} = \gamma PRM* \left({\log(i)}/{i}\right) ^ {1/d}$
   \STATE $nearest\_n = nearest(graph,new\_node,R_{PRM*})$
   \FOR{$i=1$ {\bfseries to} $number\_of\_nearest\_nodes$}
        \STATE $edge\_link = createEdge(new\_node,nearest[i])$
        \IF{$edgeNotInObstacleReigon(edge\_link)$}
            \STATE $addToGraph(edge\_link)$
        \ENDIF
   \ENDFOR
   \STATE $i = i + 1$
   \UNTIL{$i < iteration\_limit$}
\end{algorithmic}
\end{algorithm}

\subsection{RRT}
RRT (Rapidly exploring Random Trees) explores the path to the goal while creating the graph. First RRT generates random nodes and checks if this node does not fall into any obstacle before connecting it to the closest node in the graph. Edge to the closest node should also avoid the path going through any obstacle. The end condition for this algorithm is hit either we created a point inside the goal area or the time limit/number of tries has reached its maximum limit. To generate random nodes any random generator can be used as every random generator will bring its own bias so we can say that it affects the path resulting from RRT. This paper \cite{doi:10.1177/0278364911406761} talks about sampling theory in detail.
\\RRT is a fairly quick and easy-to-implement \cref{RRT*-Algorithm} but it does not guarantee optimality. It also produces cubic graphs which is solved by RRT*.

\begin{algorithm}
\caption{RRT Algorithm}
\label{RRT-Algorithm}
\begin{algorithmic}
    \STATE {\bfseries Input:} $graph, iteration\_limit$
    \REPEAT
    \STATE Initialize $i = 0$
    \STATE $new\_node = random\_sampler$
    \IF {$(notInObstacleReigon(new\_state)$)} 
        \STATE $nearest\_n = nearest(graph, new\_node)$
        \STATE $edge = createEdge(new\_node,nearest\_n)$
            \IF {$(notInObstacleReigon(edge)$)}
                \STATE $addToGraph(edge)$
            \ENDIF
        \ENDIF
    \STATE $i = i + 1$
    \UNTIL{$i < iteration\_limit$}
\end{algorithmic}
\end{algorithm}

\subsection{RRT*}
RRT* is an optimized \cite{doi:10.1177/0278364911406761} version of the Rapidly-exploring random trees algorithm. It guarantees the shortest path to be returned. In addition to the RRT algorithm, RRT* calculates the distance to k nearest nodes and then calculates the total cost to reach a new node via each neighbor. It chooses the minimum cost path and connects the neighbor and the new node. One more step is to rewire the tree to connect the lowest-cost paths. For each neighbor, the cost of traversal through the new node is checked, and if that is less, the neighbor's connection to its previous parent is destroyed and the new node becomes the parent node. RRT* \cref{RRT*-Algorithm} results in quite a straight path. It also performs better than RRT in obstacle-dense environments. It has added computation hence performance gets worse. 

\begin{algorithm}[tb]
\caption{RRT* Algorithm}
\label{RRT*-Algorithm}
\begin{algorithmic}
    \STATE {\bfseries Input:} $graph, iteration\_limit$
    \REPEAT
    \STATE Initialize $i = 0$
    \STATE $new\_node = random\_sampler$
    \IF {$(notInObstacleReigon(new\_state)$)}
        \STATE $min\_cost = \infty$
        \STATE $k\_nearest = kNearest(graph, new\_node)$
        \FOR{$i=1$ {\bfseries to} $number\_of\_nearest\_nodes$}
            \STATE $curr\_cost = cost(new\_node, k\_nearest[j]) + cost(k\_nearest[j])$
            \IF {$curr\_cost < min\_cost$}
                \STATE $e = createEdge(new\_node,k\_nearest[j])$
                \IF {$(notInObstacleReigon(e)$}
                    \STATE $nearest\_edge\_link = e$
                    \STATE $min\_cost = curr\_cost$
                \ENDIF
            \ENDIF
        \ENDFOR
        \STATE $addToGraph(e)$
        \FOR{$k=1$ {\bfseries to} $number\_of\_nearest\_nodes$}
            \IF {$new\_cost\_near[k] < prev\_cost\_near[k]$}
                \STATE $e= createEdge(new\_node, k\_nearest[k])$
                \IF {$(notInObstacleReigon(e)$}
                \STATE $addToGraph(e)$
                \STATE $parent(k\_nearest[k]) = new\_node$
                \ENDIF
            \ENDIF
        \ENDFOR
    \ENDIF
    \STATE $i = i + 1$
    \UNTIL{$i < iteration\_limit$}
\end{algorithmic}
\end{algorithm}

\section{Future Extensions}
There are various other optimization techniques added to these algorithms in the last decade e.g. Lazy PRM \cite{bohlin2000path}, Obstalc-based PRM \cite{amato1998obprm}, Informed RRT \cite{gammell2014informed}, Probabilistic Roadmap of Tree (PRT) \cite{akinc2005probabilistic}, sPRM, k-sPRM, k-PRM*, RRG, k-RRG, k-RRT* \cite{karaman2011sampling}, etc. These algorithms should be evaluated as per \cref{completeness-optimality} and \cref{time-space-complexity}.

\section{Conclusion}
RRT algorithms are good at finding a possible path from a given point, every time a new location is given an entirely new graph will be generated. Repeating generating paths can lead to inefficiency. PRM provides an alternate approach to building the graph once to cover most of the obstacle-free area. On top of it shortest path planning algorithm (A*, Dijkstra, etc.) \cite{motion_planning_survey_3} can be used to find the path from one node to another. PRM doesn't provide a graph itself, it outputs a graph with nodes and edges to represent connectivity.

Sampling-based path planner comes with their pros and cons list. For a larger configuration space, these algorithms will be fast and more efficient\cite{sampling_based_planner_1}. Sampling-based path planners can create obstacle-free feasible edge connectivity. In Sampling-based path planners, we generate and connect random points so we can reach the solution faster. Sampling-based planners can produce a sub-optimal path in a given time limit. Due to the randomness of the generating points, they tend to be random in providing the solution. It is quite possible to not get the same path generated next time.

\cref{completeness-optimality} shows the completeness and optimality of PRM, PRM*, RRT, and RRT*. \cref{time-space-complexity} represents the time and space complexity of these algorithms.

\begin{table}[tb]
\caption{Comparison based on Completeness, Optimality of various Sampling-based path planners}
\label{completeness-optimality}
\vskip 0.15in
\begin{center}
\begin{small}
\begin{sc}
\begin{tabular}{lcccr}
\hline
\abovespace\belowspace
Algorithm & Completeness & Optimality \\
\hline
\abovespace
PRM    & $\surd$ & $\times$\\
PRM*   & $\surd$ & $\surd$\\
RRT    & $\surd$ & $\times$ \\
\belowspace
RRT*   & $\surd$ & $\surd$ \\
\hline
\end{tabular}
\end{sc}
\end{small}
\end{center}
\vskip -0.1in
\end{table}

\begin{table}[tb]
\caption{Comparison based on Time and Space complexity of various Sampling-based path planners}
\label{time-space-complexity}
\vskip 0.15in
\begin{center}
\begin{small}
\begin{sc}
\begin{tabular}{lcccr}
\hline
\abovespace\belowspace
Algorithm & Time Complexity & Space Complexity \\
\hline
\abovespace
PRM    & $O(n\log n)$ & $O(n)$\\
PRM*   & $O(n\log n)$ & $O(n\log n)$\\
RRT    & $O(n\log n)$ & $O(n)$\\
\belowspace
RRT*   & $O(n\log n)$ & $O(n)$\\
\hline
\end{tabular}
\end{sc}
\end{small}
\end{center}
\vskip -0.1in
\end{table}

\bibliography{example_paper}
\bibliographystyle{icml2023}

%%%%%%%%%%%%%%%%%%%%%%%%%%%%%%%%%%%%%%%%%%%%%%%%%%%%%%%%%%%%%%%%%%%%%%%%%%%%%%%
%%%%%%%%%%%%%%%%%%%%%%%%%%%%%%%%%%%%%%%%%%%%%%%%%%%%%%%%%%%%%%%%%%%%%%%%%%%%%%%
% APPENDIX
%%%%%%%%%%%%%%%%%%%%%%%%%%%%%%%%%%%%%%%%%%%%%%%%%%%%%%%%%%%%%%%%%%%%%%%%%%%%%%%
%%%%%%%%%%%%%%%%%%%%%%%%%%%%%%%%%%%%%%%%%%%%%%%%%%%%%%%%%%%%%%%%%%%%%%%%%%%%%%%
% \newpage
% \appendix
% \onecolumn

% You can have as much text here as you want. The main body must be at most $8$ pages long.
% For the final version, one more page can be added.
% If you want, you can use an appendix like this one, even using the one-column format.
%%%%%%%%%%%%%%%%%%%%%%%%%%%%%%%%%%%%%%%%%%%%%%%%%%%%%%%%%%%%%%%%%%%%%%%%%%%%%%%
%%%%%%%%%%%%%%%%%%%%%%%%%%%%%%%%%%%%%%%%%%%%%%%%%%%%%%%%%%%%%%%%%%%%%%%%%%%%%%%

\end{document}